\documentclass[9pt,twocolumn,twoside]{osajnl}

\usepackage{amsmath}
\usepackage{bm}
\usepackage{color}
\usepackage{graphicx}

\newcommand{\prop}[2]{\mathcal{{\rm Prop}}_{#1} \bigl[#2\bigl]}

\newcommand{\fig}[1]{Fig.\ref{#1}}

\newcommand{\Fig}[1]{Figure \ref{#1}}

\journal{ao} % Choose journal (ao, ol, josaa, josab)

\setboolean{shortarticle}{false} % true = letter, false = research article

\title{Deep-learning-based data page classification for holographic memory}

\author[1,*]{Tomoyoshi Shimobaba}
\author[1]{Naoki Kuwata}
\author[2]{Mizuha Homma}
\author[1]{Takayuki Takahashi}
\author[1]{Yuki Nagahama}
\author[1]{Marie Sano}
\author[1]{Satoki Hasegawa}
\author[1]{Ryuji Hirayama}
\author[1]{Takashi Kakue}
\author[3]{Atsushi Shiraki}
\author[4]{Naoki Takada}
\author[1]{Tomoyoshi Ito}

\affil[1]{Graduate School of Engineering, Chiba University, 1-33 Yayoi-cho, Inage-ku, Chiba, 263-8522, Japan}
\affil[2]{Department of Electrical and Electronics Engineering, Chiba University, 1-33 Yayoi-cho, Inage-ku, Chiba, 263-8522, Japan}
\affil[3]{Institute of Management and Information Technologies, Chiba University, 1-33 Yayoi-cho, Inage-ku, Chiba, 263-8522, Japan}
\affil[4]{Science Department, Natural Sciences Cluster, Research and Education Faculty, Kochi University, Kochi 780-8520, Japan}

\affil[*]{Corresponding author: shimobaba@faculty.chiba-u.jp}

\dates{Compiled \today}

\ociscodes{(090.0090) Holography;  (090.1760) Computer holography; (090.2900) Optical storage materials.}

\doi{\url{http://dx.doi.org/10.1364/ao.XX.XXXXXX}}

\begin{abstract}
We propose a deep-learning-based classification of data pages used in holographic memory. 
We numerically investigated the classification performance of a conventional multi-layer perceptron (MLP) and a deep neural network, under the condition that reconstructed page data are contaminated by some noise and are randomly laterally shifted.
The MLP was found to have a classification accuracy of 91.58\%, whereas the deep neural network was able to classify data pages at an accuracy of 99.98\%.
The accuracy of the deep neural network is two orders of magnitude better than the MLP.
\end{abstract}

\setboolean{displaycopyright}{true}

\begin{document}

\maketitle
\thispagestyle{fancy}
\ifthenelse{\boolean{shortarticle}}{\abscontent}{}

\section{Introduction}
Optical memories such as compact discs (CD), digital versatile discs (DVD) and Blu-ray discs utilize laser spots to read and write digital data. The recording disk is irradiated with the laser spot, condensed by a lens. 
Increasing the memory capacity requires miniaturization of  the laser spot.
In Blu-ray discs, the minimum spot is approximately 0.15$\mu$m.
To achieve higher density, it is necessary to use a laser with a shorter wavelength and a lens with a higher numerical aperture, but it is coming to its limit.

In holographic memory \cite{psaltis1998holographic,Ruan2014Recent}, digital data is converted into a two-dimensional pattern called a data page, and this is recorded on a recording medium as a hologram. 
The main features of the holographic memory are below:  (1) the access speed is fast because the data page can be read and written as two-dimensional images, and (2) multiple data pages can be stored in the same recording area by multiplex recording characteristics of holography, leading to an increase in the memory capacity.
In addition, the combination of holographic memory and optical encryption is interesting because the increased security, does not come at the expense of increased encryption and decryption time \cite{Matoba1999Encrypted} .

Although holographic memory has a significant advantage as the next-generation data storage, problems do exist, such as bit errors arising from pixel misalignment and noise. Simple thresholding to a data page detected by an image sensor induces bit errors; therefore, we need sophisticated methods to correctly detect bits.
Precise pixel alignment between reconstructed data pages and an imaging device are required to correctly read data pages. Solutions to this issue have been proposed \cite{Burr2001Compensation}.
Some noises (speckle noise, interpixel interference, interpage interference \cite{Ruan2014Recent}) contaminate reconstructed  data page. Solutions to the noises have been proposed, such as a  Viterbi algorithm \cite{Heanue1996Signal}, a deconvolution method \cite{Lee2011Increasing}, a gradient decent method \cite{Kim2014Iterative} and an autoencoder \cite{shimobaba2017autoencoder}.
%In addition, although sacrificing the coding efficiency somewhat, data pages coded via modulation and error-correction codes helps to significantly improve bit errors compared to raw data page
In addition, data pages coded via modulation and error-correction codes help in significantly improving bit errors when compared to raw data pages \cite{Burr1997Modulation}.

In this study, we propose a deep-learning-based classification of data pages. 
The deep neural network (convolutional neural network \cite{Krizhevsky2012Imagenet}) is learned, which is composed of convolutional layers, pooling layer, and a fully-connected layers, using data pages reconstructed from holograms, and then the learned deep neural network classified data pages. 
It is significant that the deep neural network can automatically acquire optimum data page classifications from learning datasets without human intervention.
%This feature means that it is not necessary to redesign the algorithm even if holographic memory system is changed.

We numerically investigated the classification performance between a conventional multi-layer perceptron (MLP) and the deep neural network. Under the condition that  reconstructed data pages are contaminated by some noise and are randomly laterally shifted, the MLP had a classification accuracy of only 91.58\%, whereas the deep neural network was able to classify the data pages at an accuracy of 99.98\%.

\section{Deep-learning-based classification for holographic memory}
\begin{figure}[htbp]
\centering
\fbox{\includegraphics[width=\linewidth]{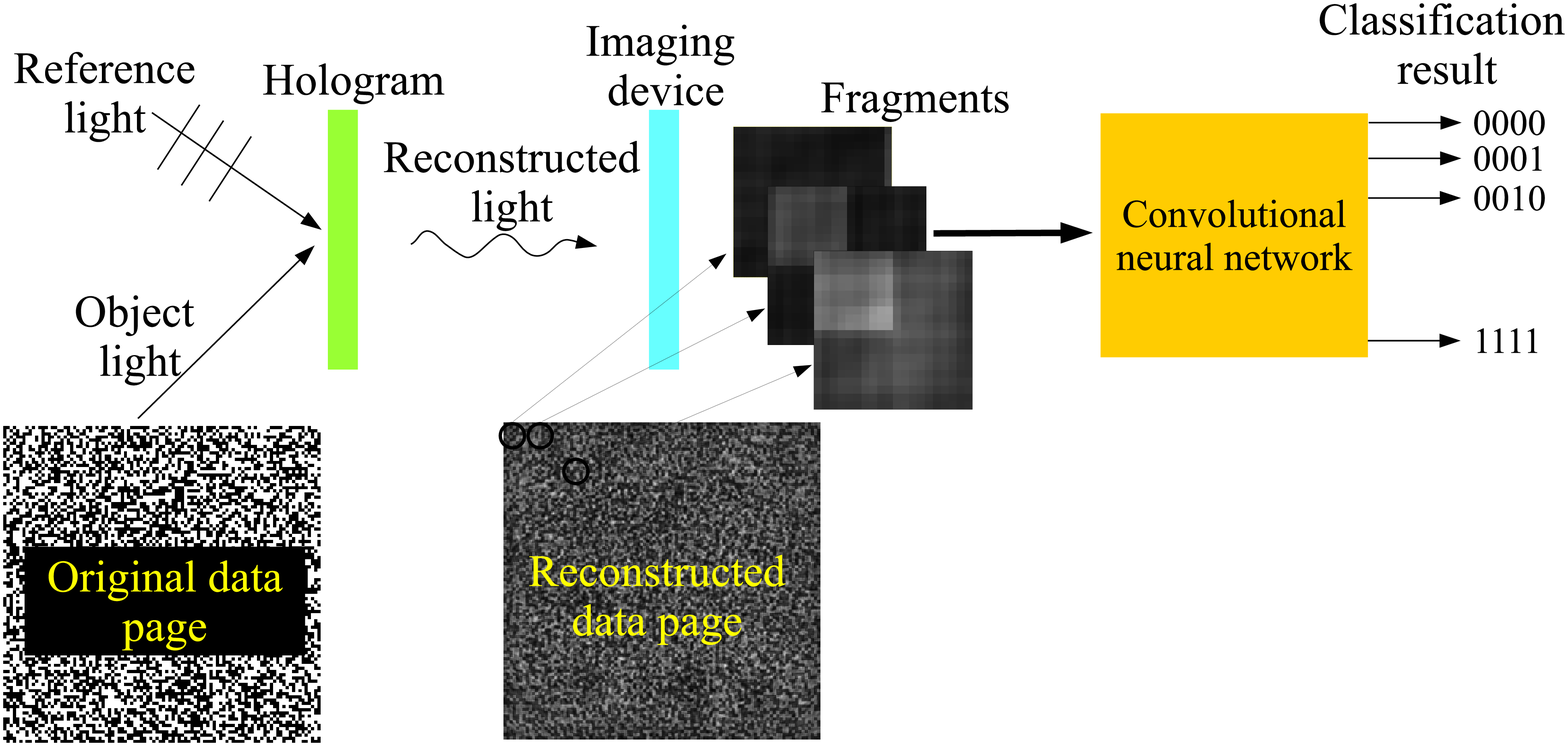}}
\caption{Deep learning-based classification for holographic memory.}
\label{fig:system}
\end{figure}

\begin{figure}[htbp]
\centering
%\fbox{\includegraphics[width=\linewidth]{fig_system}}
\fbox{\includegraphics[width=\linewidth]{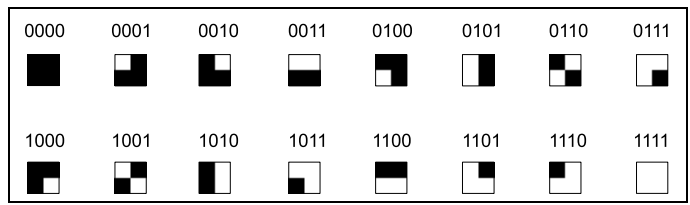}}
\caption{4-bit representation using 16 fragments.}
\label{fig:binary_pattern}
\end{figure}

Figure \ref{fig:system} shows the deep-learning-base classification for holographic memory, where data pages are recorded as holograms. 
These data pages are composed of 4-bit patterns as shown in \fig{fig:binary_pattern}.
In this paper, we do not use multiple recording of data pages in the same region of the hologram, or modulation codes such as 6:8 modulation code, or error-correction codes.
Imaging devices such as charge coupled devices (CCDs) and complementary MOS (CMOS) cameras captured the reconstructed data pages, which are contaminated by noise, from the holograms.
The reconstructed data pages are divided into fragments corresponding to 4-bit original data.
The deep neural network in which we used a convolutional neural network (CNN) 
classifies fragments into the most similar 4-bit pattern.

\subsection{Hologram generation}
In this study, we used amplitude holograms $I({\bm m})$ that are generated from data pages. The amplitude holograms are generated by
\begin{equation}
I({\bm m}) = |O({\bm m}) + R({\bm m})|^2,
\end{equation}
where ${\bm m}$ denotes a two-dimensional position vector in the hologram plane, $O({\bm m})$ is the object light of data pages that are displayed on a spatial light modulator, and $R({\bm m})$ is the reference light.
The object light is obtained from the original data page, $u({\bm n})$, using 
\begin{equation}
O({\bm m}) = \prop{z}{u({\bm n}))},
\label{eqn:fre2}
\end{equation}
where $\prop{z}{\cdot}$ denotes the diffraction operator with a propagation distance of $z$.
%, and $p(\bm{n})$ is a pseudo-random number ranging from 0 to 1, which is used for expressing the random phase plates.
${\bm n}$ denotes a two-dimensional position vector in the object plane.
The reconstructed data pages, $u'(\bm{n})$, from the holograms are obtained by
\begin{equation}
u`(\bm{n}) = |\prop{-z}{I(\bm{m})}|^2.
\label{eqn:fre2}
\end{equation}
In this study, $R({\bm m})$ is the inline planar wave, so that the mathematical expression is simply $R({\bm m})=1$.
Thus, the reconstructed data pages were degraded by direct light and conjugate light.

\subsection{Convolutional neural network}
\begin{figure}[htbp]
\centering
\fbox{\includegraphics[width=\linewidth]{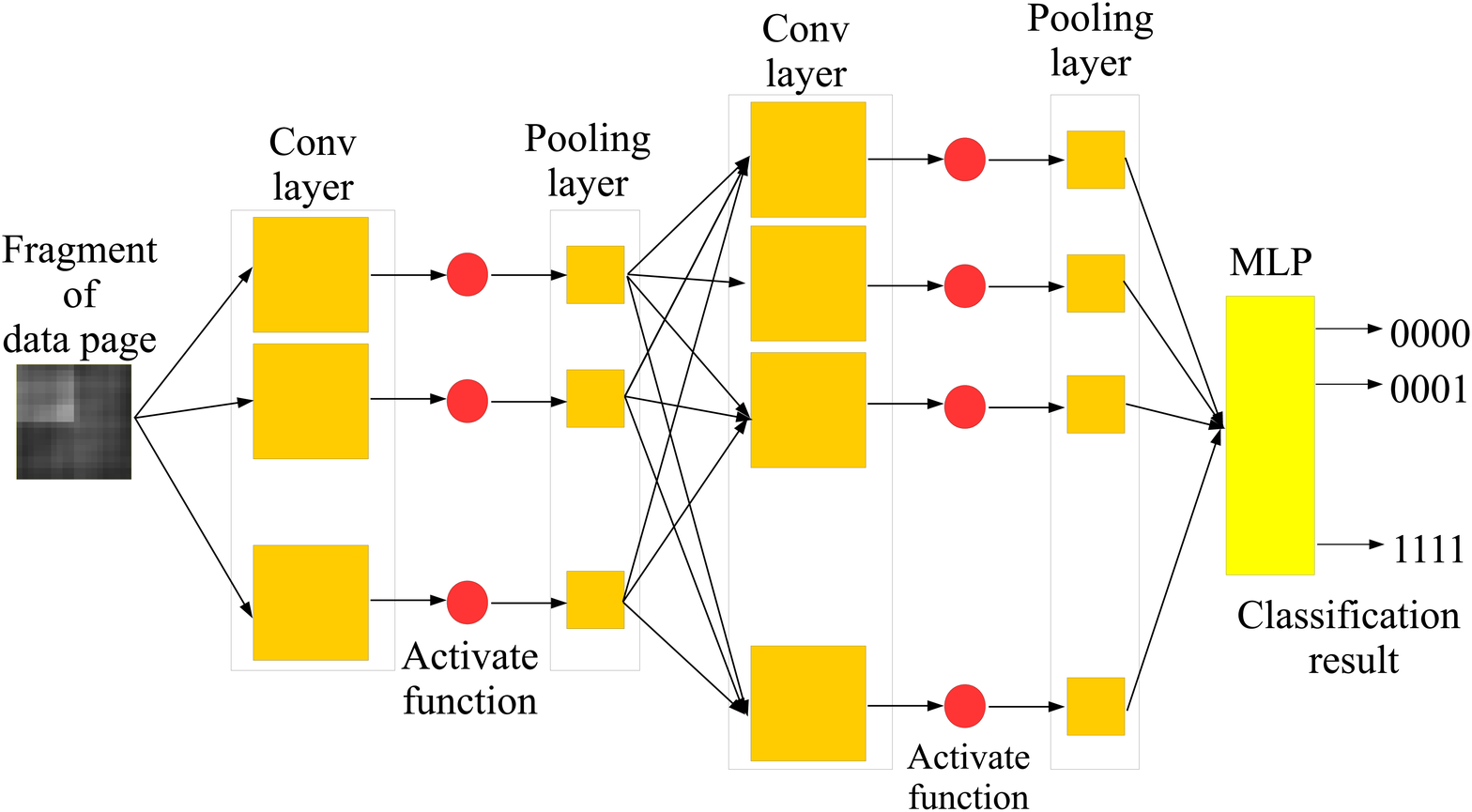}}
\caption{Deep-learning-based classification for holographic memory.}
\label{fig:cnn}
\end{figure}

\Fig{fig:cnn} shows a CNN for classifying fragments of data pages.
The CNN consists of convolution layers, pooling layers, and an MLP composed of fully connected layers and an output layer.

A convolution layer automatically acquires feature maps of input two-dimensional (2D) data $x_{ij}$, where the subscript $ij$ denote the pixel index, using $M$ filters whose filter coefficients denote $h^{(m)}_{pq}$ where $m \in [0,M-1]$.
When setting $M$ different filters, the convolution layer acquire $M$ different feature maps.
The output of the layer $y^{(m)}_{ij}$ is calculated by
\begin{equation}
y^{(m)}_{ij}=f\left(\sum_{p=0}^{H} \sum_{q=0}^{H} h^{(m)}_{pq} x_{ij} + b_{ij}\right)
\end{equation}
where $f(\cdot)$ is an activate function, $H$ is the filter size and $b_{ij}$ is a bias.
We used Leaky ReLU function ($f(x)=x$ when $x>0$; otherwise, $f(x)=0.01x$) as the activate function because we confirmed that the classification performance of the activate function was better than that of the ReLU function for our situation.
We used $H=5$ in the first convolutional layer filter and $H=3$ in the second convolutional layer.

A pooling layer had the effect of reducing the sensitivity of lateral movement of the input data.
 In addition, this layer was used for reducing the input data size, resulting in a decrease of the computational complexity. 
Several pooling layers have been proposed.
The max pooling layer that we used was calculated by
\begin{equation}
y_{ij}={\rm max}\{x_{ij}\}.
\end{equation}
Here, this layer divides the input 2D data $x_{ij}$ with $W \times W$ pixels into  sub 2D images, and the maximum values in the sub images are selected and are used to generate the output image $y_{ij}$ with $W/2 \times W/2$ pixels.

\begin{figure}[htbp]
\centering
\fbox{\includegraphics[width=\linewidth]{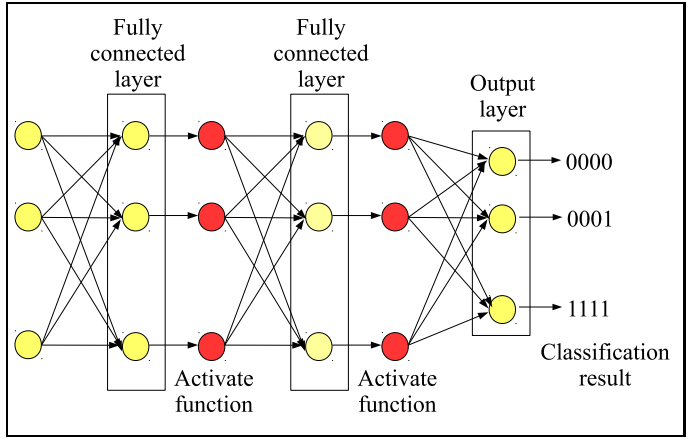}}
\caption{Multiple layer perceptron (MLP).}
\label{fig:mlp}
\end{figure}

The MLP classifies fragments into the most similar 4-bit patterns, as  shown in \fig{fig:binary_pattern}.
The structure of the MLP is shown in \fig{fig:mlp}.
A fully connected layer in the MLP was calculated by
\begin{equation}
y_{j}=f\left( \sum_{i=1}^{U} w_{ji} x_i + b_j \right),
\end{equation}
where $f(\cdot)$ is an activate function, $U$ is the number of  units, $x_i$ is the input data, $w_{ji}$ is the weight coefficients, $y_{j}$ is the output data, and $b_j$ is a bias.

In the first and second fully connected layers, we used $U=128$ and the ReLU function as the activate function.
The output layer calculates probabilities of classification using the softmax function, expressed as
\begin{equation}
y_j=\frac{\exp(x_j)}{ \sum_{i=1}^{U} \exp(x_i)  },
\end{equation}
where $U=16$ because we wanted to classify the 16 fragments shown in \fig{fig:binary_pattern}.

In the learning process of the CNN, we needed to prepare a large number of datasets composed of reconstructed fragments and corresponding correct answers. 
Using these datasets, we optimized the parameters (the filter coefficients $h^{(m)}_{pq}$, weight coefficients $w_{ji}$, and biases) in the CNN.
These parameters were optimized by minimizing a cost function.
In this study, we used the cross-entropy cost function, and, we used Adam \cite{Kingma2014Adam}, which is a stochastic gradient descent (SGD) method, as the optimizer, to minimize the cross-entropy cost function. 
This SGD randomly selects $B$ datasets among all of the datasets.
$B$ is referred to as the batch size, and here, we used a batch size of 100.
In addition, we used the Dropout method \cite{Srivastava2014Dropout} to prevent overfitting in the CNN.
Dropout randomly disables $d$ percent of units during the training process. We used $d=0.25$\% in the pooling layers and  $d=0.5$\% in the fully connected layers. 
%During each iteration step in the SGD, $N_d$ units are disabled units are randomly changed.

\section{Results}
We compared the classification performance between the CNN, show in \fig{fig:cnn}, and a conventional MLP, shown in \fig{fig:mlp}.

The data pages and holographic reconstructions had $1,000 \times 1,000$ pixels, and the fragments had $20 \times 20$ pixels. The on-bit (1) and off-bit (0) in a fragment were both expressed by $10 \times 10$ pixels.
Thus, the number of fragments per one data page and holographic reconstruction was 2,500.

We prepared 375,000 fragments (150 data pages $\times$ 2,500 fragments) and their holographic reconstructions for the training of the CNN and MLP.
For the testing of the CNN and MLP, we used another 125,000 fragments (50 data pages $\times$ 2,500 fragments) and their holographic reconstructions.
The condition for the hologram calculation were are a wavelength of 633nm and sampling pitches of the holograms and reconstructions of 4$\mu$m. 
We used the angular spectrum method for the diffraction calculation.

\begin{figure}[htbp]
\centering
\fbox{\includegraphics[width=\linewidth]{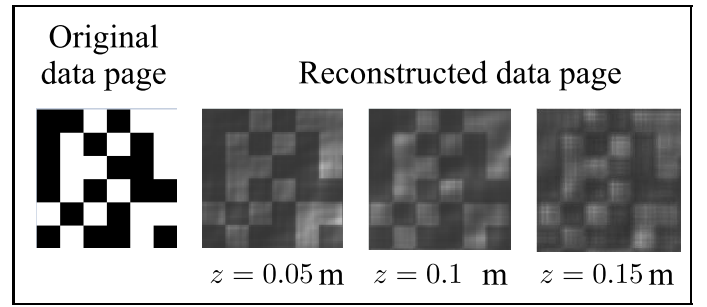}}
\caption{Examples of reconstructed data pages when changing the propagation distance $z$.}
\label{fig:reconst}
\end{figure}

We verified the classification performance of the CNN and the conventional MLP when changing the propagation distance $z$ in the hologram generation.
\Fig{fig:reconst} shows a part of the reconstructed data pages when changing the propagation distance $z$. We used $z$=0.05, 0.1 and 0.15 m.
As seen, the reconstructed data pages became blurred as the propagation distance increased, resulting in increased difficulty in the classification at the longer distance.
In addition, we added the Gaussian noise with a mean of 0 and a standard variation of 2.5 to the intensity of the reconstructed data pages to verify the robustness against intensity noise.
We added a random lateral shift of $\pm 5$ pixels to the reconstructed data pages to verify the robustness against the misalignment between the reconstructed data pages and the image sensor.

Table \ref{label:accuracy} shows the accuracy of the classification of the MLP and CNN when changing the propagation distance. 
Generally, for the accuracy metrics, in general, a bit error rate (BER) is used in holographic memory; however, we used a fragment error rate (FER) instead of the BER because the MLP and CNN classify the fragments.
The FER was calculated by ${\rm FER}=N_e / N_t$, where $N_e$ is the number of error fragments and $N_t$ is the number of total fragments in the test process.
The FER of the MLP was around $10^{-2}$, except at $z=$0.15m, whereas the FER of the CNN was around $10^{-4}$, except at $z=$0.15 m, even if the propagation distance was changed.
The CNN has an accuracy two orders of magnitude better than the MLP.

All calculations were done by deep-learning framework \cite{keras} and our wave optics library \cite{shimobaba2012computational}.

\begin{table}[]
\centering
\caption{Fragment error rate (FER) when changing the propagation distance.}
\label{label:accuracy}
\begin{tabular}{|c|c|c|l}
\cline{1-3}
      & \multicolumn{2}{c|}{Fragment error rate} &  \\ \cline{1-3}
z (m) & MLP                 & CNN  (proposal)              &  \\ \cline{1-3}
0.05  & $8.42 \times 10^{-2}$  & $1.52\times10^{-4}$         &  \\ \cline{1-3}
0.1   &  $5.07\times10^{-2}$    &  $5.84\times10^{-4}$              &  \\ \cline{1-3}
0.15  &  $1.88\times10^{-1}$   & $2.45\times10^{-2}$              &  \\ \cline{1-3}
\end{tabular}
\end{table}

\section{Conclusion}
We proposed a CNN-based data page classification for holographic memory and compared the classification performance between a conventional MLP and a CNN. 
Even if the reconstructed data pages were contaminated by some noise and were randomly laterally shifted, the CNN could classify the data pages in fragment error rates of around $10^{-4}$.
It is significant that the deep neural network can automatically acquire optimum data page classification from learning data without human intervention. %This feature means that it is not necessary to redesign the algorithm even if holographic memory system is changed.
In this study, we used raw data pages without any modulation codes or error-correction codes. If we use these codes in the CNN, we expect to increase the classification performance.
In our upcoming project, we plan to verify the CNN performance with these coding methods in a more realistic environment simulation \cite{Kinoshita2005Integrated}.

\section*{Funding}
This work was partially supported by JSPS KAKENHI Grant Numbers 16K00151.

%\section*{Appendix}

% Bibliography
%\bibliography{sample}

% Full bibliography added automatically for Optics Letters submissions
% Note that this extra page will not count against page length
\ifthenelse{\boolean{shortarticle}}{%
\clearpage
\bibliographyfullrefs{sample}
}{}

%Manual citation list
%\begin{thebibliography}{1}
%\bibitem{Zhang:14}
%Y.~Zhang, S.~Qiao, L.~Sun, Q.~W. Shi, W.~Huang, %L.~Li, and Z.~Yang,
% \enquote{Photoinduced active terahertz metamaterials with nanostructured
%vanadium dioxide film deposited by sol-gel method,} Opt. Express \textbf{22},
%11070--11078 (2014).
%\end{thebibliography}

\end{document}